 \documentclass[pmlr,twocolumn,10pt]{jmlr} 

\usepackage{caption}
\usepackage{array}
\usepackage{float}
\usepackage{booktabs}

\jmlrvolume{LEAVE UNSET}
\jmlryear{2023}
\jmlrsubmitted{LEAVE UNSET}
\jmlrpublished{LEAVE UNSET}
\jmlrworkshop{Machine Learning for Health (ML4H) 2023} 

\title[Causal prediction models for VAKI diagnosis]{Causal prediction models for medication safety monitoring: The diagnosis of vancomycin-induced acute kidney injury}

\author{%
\Name{Izak {Yasrebi-de Kom}}\textsuperscript{1,3}\thanks{Corresponding author} \Email{i.a.r.dekom@amsterdamumc.nl} \\
\Name{Joanna {Klopotowska}}\textsuperscript{1,3} \Email{j.e.klopotowska@amsterdamumc.nl} \\
\Name{Dave {Dongelmans}}\textsuperscript{2,3} \Email{d.a.dongelmans@amsterdamumc.nl} \\
\Name{Nicolette {De Keizer}}\textsuperscript{1,3} \Email{n.f.keizer@amsterdamumc.nl} \\
\Name{Kitty {Jager}}\textsuperscript{1,3} 
\Email{k.j.jager@amsterdamumc.nl} \\
\Name{Ameen {Abu-Hanna}}\textsuperscript{1,3}
\Email{a.abu-hanna@amsterdamumc.nl} \\
\Name{Giovanni {Cinà}}\textsuperscript{1,3,4,5}
\Email{g.cina@amsterdamumc.nl} \\
\addr \textsuperscript{1}Amsterdam UMC location University of Amsterdam, Department of Medical Informatics, The Netherlands \\
\addr \textsuperscript{2}Amsterdam UMC location University of Amsterdam, Department of Intensive Care Medicine, The Netherlands \\
\addr \textsuperscript{3}Amsterdam Public Health, The Netherlands \\
\addr \textsuperscript{4}Institute for Logic, Language and Computation, University of Amsterdam, The Netherlands \\
\addr \textsuperscript{5}Pacmed, The Netherlands
\vspace{-1cm}
}

\begin{document}

\maketitle

\begin{abstract}
The current best practice approach for the retrospective diagnosis of adverse drug events (ADEs) in hospitalized patients relies on a full patient chart review and a formal causality assessment by multiple medical experts. This evaluation serves to \textit{qualitatively} estimate the probability of causation (PC); the probability that a drug was a necessary cause of an adverse event. This practice is manual, resource intensive and prone to human biases, and may thus benefit from data-driven decision support. Here, we pioneer a causal modeling approach using observational data to estimate a lower bound of the PC (PC\textsubscript{low}). This method includes two key causal inference components: (1) the target trial emulation framework and (2) estimation of individualized treatment effects using machine learning. We apply our method to the clinically relevant use-case of vancomycin-induced acute kidney injury in intensive care patients, and compare our causal model-based PC\textsubscript{low} estimates to qualitative estimates of the PC provided by a medical expert. Important limitations and potential improvements are discussed, and we conclude that future improved causal models could provide essential data-driven support for medication safety monitoring in hospitalized patients.

\end{abstract}
\vspace{2cm}

\section{Introduction}
\label{sec:intro}

Adverse drug events (ADEs) are common in hospitalized patients, affecting up to three in ten patients \citep{bates1995incidence, klopotowska2013adverse}. Although the diagnosis of ADEs (i.e., detection \textit{after the fact}) is of vital importance for patient safety and outcomes, ADEs often remain undetected \citep{bates2023safety, klopotowska2013recognition}. The current best practice approach for ADE diagnosis relies on a full patient chart review and a formal causality assessment by multiple medical experts to \textit{qualitatively} estimate the probability of causation (PC); the probability that a drug was a necessary cause of an adverse event \citep{klopotowska2013adverse, WHO_causality, naranjo1981method}. This approach is manual and resource intensive, thereby hampering large scale medication safety monitoring and its use in daily, often busy, clinical practice. Furthermore, this approach suffers from low inter-rater reliability and a high risk of bias \citep{varallo2017imputation}. Novel methods for the diagnosis of ADEs are therefore urgently needed, and artificial intelligence may aid through data-driven approaches \citep{bates2023safety}. However, a recent systematic review on prediction models for the diagnosis (and prognosis) of ADEs in hospitalized patients noticed a lack of studies addressing the causal aspect of ADEs in their modeling approach \citep{yasrebi2023electronic}. Importantly, addressing this causal aspect could alleviate the above mentioned biases from human assessors.

The PC answers the following counterfactual question: \textit{Given a patient that received a drug and subsequently developed an adverse event, what is the probability that the patient would not have developed the adverse event if the patient had not received the drug?} Past seminal work uncovered that the PC is non-identifiable from epidemiological data \citep{robins1989probability, greenland1988conceptual}. Nonetheless, an informative lower bound on the PC (PC\textsubscript{low}) {\it is} identifiable given strong assumptions \citep{robins1989probability, suzuki2012relations, tian2000probabilities}. 

Here, we pioneer a data-driven method to estimate the PC\textsubscript{low} using causal prediction models, which may support the current best practice approach. To the best of our knowledge, the PC\textsubscript{low} is a novel estimand in the setting of causal model-based ADE diagnosis. Our method combines two important causal inference concepts: the target trial emulation (TTE) framework \citep{hernan2016using} and \textit{individualized} treatment effect (ITE) estimation using machine learning. The TTE framework serves to support causal inference using observational data through the design of a hypothetical randomized controlled trial. This process aids in the identification and avoidance of biases and limitations related to the approach and the data. Estimation of ITEs using machine learning shows promising results \citep{bica2021real, kunzel2019metalearners}. Given that we expect the adverse effects of drugs to vary across individuals \citep{kim2022risk}, obtaining individualized PC\textsubscript{low} estimates seems to be necessary for ADE diagnosis.

We apply our method in the clinically relevant and frequently occurring use-case of vancomycin-induced acute kidney injury (VAKI) in intensive care unit (ICU) patients \citep{kan2022vancomycin}. Vancomycin - an antibiotic - can be \textit{nephrotoxic} (i.e. toxic for the kidneys), and thereby cause a rapid decline in kidney function, also known as acute kidney injury (AKI) \citep{kellum2012kidney}. Although we use VAKI as our use-case, the presented approach may be applicable for any other ADE. 

\section{Methods}
\label{sec:methods}

\subsection{The Probability of Causation}
\label{sec:pc}

Previous work uncovered the assumptions needed to estimate the PC\textsubscript{low} \citep{robins1989probability, suzuki2012relations, tian2000probabilities}. These assumptions largely overlap with those commonly described in causal inference literature regarding the estimation of average treatment effects on the treated: the stable unit treatment value assumption \citep{vanderweele2013causal, rubin1980randomization}, weak positivity \citep{imbens2004nonparametric} and conditional ignorability for controls \citep{imbens2004nonparametric, heckman1997matching}. However, additional strong assumptions are needed concerning the biological mechanisms that connect the treatment with the outcome, censoring events and competing risks: a monotonic relationship between the treatment and the outcome conditional on the covariates, and a non-preventive relationship between the treatment and censoring events and competing risks \citep{robins1989probability, suzuki2012relations} (\appendixref{apd:first}).

Our retrospective observational data consists of ICU admissions, each with covariates $\vec{X} \in \set{X}$, a treatment indicator $\vec{A} \in \{0,1\}$ (whose values denote an alternative antibiotic and vancomycin, respectively) and an outcome indicator $\vec{Y} \in \{0,1\}$ (denoting no AKI and AKI, respectively).

Given the above assumptions, the PC\textsubscript{low} is given by the excess-risk-ratio (ERR) when it yields a positive value \citep{robins1989probability, tian2000probabilities, suzuki2012relations}:
\begin{align}
\vec{PC_{low}}&=\max\{0, \vec{ERR}\}
\end{align}

where the ERR is given by one minus the inverse of the risk ratio (RR):

\begin{align}
\vec{ERR}&=1-\frac{1}{\vec{RR}}\nonumber\\
&=1-\frac{1}{\frac{\mathbb{E}(\vec{Y}|\vec{A}=1,\vec{X}=x)}{\mathbb{E}(\vec{Y}|\vec{A}=0,\vec{X}=x)}}\nonumber\\
&=1-\frac{\mathbb{E}(\vec{Y}|\vec{A}=0,\vec{X}=x)}{\mathbb{E}(\vec{Y}|\vec{A}=1,\vec{X}=x)}
\end{align}

Henceforth, we denote $\mathbb{E}(\vec{Y}|\vec{A}=0,\vec{X}=x)$ and $\mathbb{E}(\vec{Y}|\vec{A}=1,\vec{X}=x)$ with $\mu_0$ and $\mu_1$, respectively.

\subsection{Target Trial Emulation}
\label{sec:tte}

Although the TTE framework has mostly been applied for the estimation of average treatment effects, e.g. by \citet{hoffman2022comparison} and \citet{de2020antimicrobial}, we deem the framework to be equally useful and relevant in rigorously defining the relevant population and control group needed for estimating the PC\textsubscript{low}. We define a hypothetical trial comparing the risk of AKI in the ICU between eligible ICU admissions initiated on vancomycin and eligible ICU admissions initiated on one of several alternative, minimally nephrotoxic antibiotics with overlapping indications. We specify the trial protocol components: eligibility criteria, treatment strategies, assignment procedure, follow-up period, outcome, causal contrast of interest and analysis plan \citep{hernan2016using, fu2023target}. For each of the components, we identify limitations associated with our data and approach, and attempt to mitigate these through our trial emulation. See \appendixref{apd:second} for our TTE protocol.

\subsection{Data}
\label{sec:data}

We utilize a retrospective observational data set consisting of electronic health record (EHR) data of 176,489 consecutive admissions to 15 Dutch ICUs between January 1, 2010 and December 31, 2019. The data consist of two linked parts: (1) The Dutch National Intensive Care Evaluation (NICE) quality registry \citep{van2015data} and (2) data directly collected from the 15 ICUs. See \appendixref{apd:third} for a more elaborate description of our data.

\subsection{Modeling}
\label{sec:modeling}

We estimate $\mu_0$ and $\mu_1$ using T-learners, which are a type of meta-learners commonly applied for ITE estimation \citep{kunzel2019metalearners}. This approach involves fitting two base models that predict the outcome (AKI): one using the ICU admissions that received an alternative antibiotic ($\hat{\mu}$$_0$), and one using the ICU admissions that received vancomycin ($\hat{\mu}$$_1$). Next, these base models are applied to estimate $\mu_0$ and $\mu_1$ for \textit{all} ICU admissions, thereby estimating the counterfactuals. We develop three T-learners, each with different base models: Bayesian additive regression trees (BART) \citep{chipman2010bart}, random forests (RF) \citep{breiman2001random, obrien2019random} and logistic regression (LR). We choose the BART and RF base learners as they are commonly used and compared in ITE estimation tasks using T-learners \citep{kunzel2019metalearners, caron2022estimating}, and add the LR base learner (without interaction terms) to investigate the relative performance of a `simple' linear approach.

Using the T-learner $\mu_0$ and $\mu_1$ estimates, we calculate the PC\textsubscript{low} for all ICU admissions that received vancomycin and subsequently developed AKI. Additionally, we calculate the average treatment effect on the treated (ATT) by averaging the $\mu_0$ and $\mu_1$ estimates and subsequently calculating the absolute risk difference (ARD) and RR for the ICU admissions that received vancomycin.

We compare these ATTs to estimates obtained through inverse probability of treatment weighting (IPTW), for which we calculate the propensity scores with three different methods: BART, RF and LR. To ensure that the positivity assumption is not violated, we assess the overlap of the treatment probabilities between the treatment groups and exclude the ICU admissions outside the area of common support \citep{bergstra2019three}. As this assumption is equally important for the T-learners, we also apply this selection \textit{before} fitting these. See \appendixref{apd:third} for additional details about our methods.

\begin{figure*}[tp]
\floatconts
  {fig:pc}
  {\caption{PC\textsubscript{low} estimates of the BART, LR and RF T-learner models for admissions that received vancomycin and developed AKI (in the subset of the \textit{train} set within the area of common support, n = 131)}}
  {\includegraphics[scale=0.55]{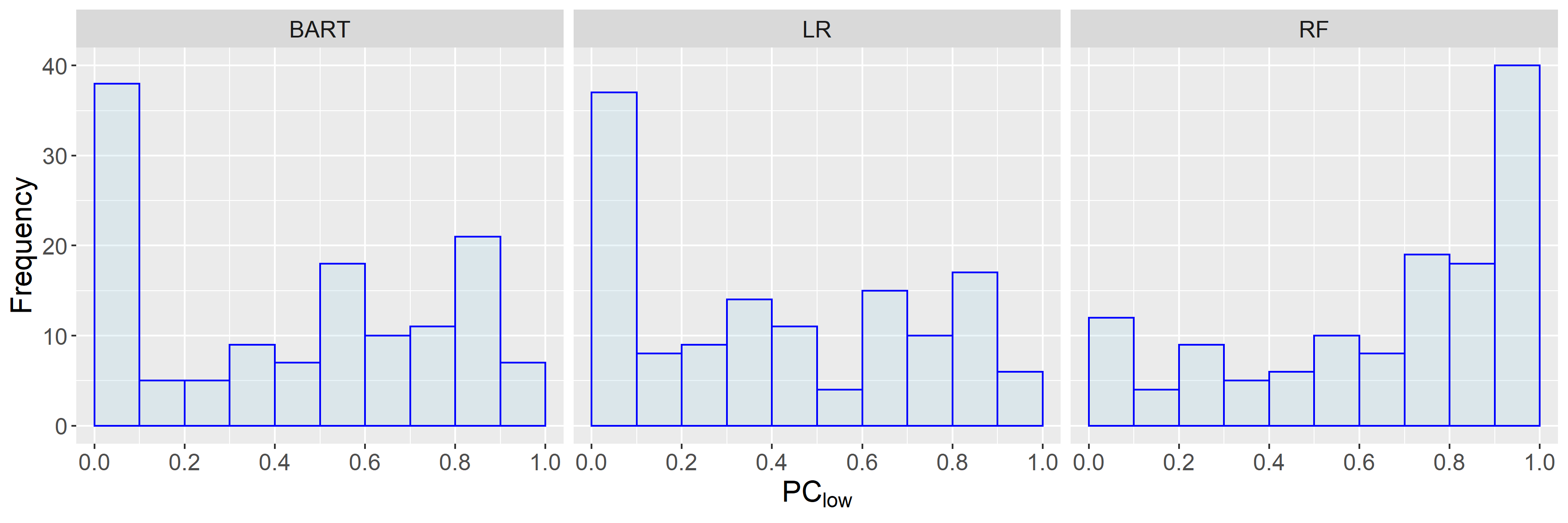}}
\end{figure*}

\subsection{Evaluation}
\label{sec:eval}

We randomly split our data into a train set (90\%) and a test set (10\%). All models are developed using the train set. We evaluate both the factual AKI and the counterfactual PC\textsubscript{low} predictions of the T-learners using the test set. For the evaluation of the factual AKI predictions, we calculate the area under the receiver operating characteristic curve (AUC) of the T-learner base models on their respective treatment arm. We evaluate the T-learners' counterfactual PC\textsubscript{low} predictions using the test set's ICU admissions that received vancomycin and subsequently developed AKI. For these admissions, one medical expert conducted formal causality assessments using the WHO-UMC criteria tailored to drug-induced AKI \citep{WHO_causality, mehta2015phenotype}. These assessments result in a qualitative estimate of the PC by labeling each case with an {\it unassessable}, {\it unlikely}, {\it possible}, {\it probable} or {\it nearly certain} causal relationship between vancomycin and AKI. We assign the following probabilities to the latter categories: 0.5, 0.25, 0.5, 0.75 and 0.9, respectively, and calculate the mean squared error (MSE). A probability of 0.5 is assigned to the {\it unassessable} category in line with a recent systematic review that reported the attributable fraction for patients treated with vancomycin and a subsequent AKI \citep{ray2016vancomycin}. Next, we dichotomize the qualitative estimates to a value of 1 for an {\it unassessible}, {\it possible}, {\it probable} and {\it nearly certain}, and a value of 0 for an {\it unlikely} causal relationship, and calculate the AUC and the positive predictive value (PPV). For the latter, we dichotomize each predicted PC\textsubscript{low} with a threshold of 0.5, as this threshold represents the idea that it was \textit{more likely than not} that vancomycin was a necessary cause of the AKI. Lastly, we calculate the correlation between the PC\textsubscript{low} predictions of the three T-learners to inspect the agreement between the predictions of the models.

We obtain 95\% confidence intervals (CIs) for the ATT and evaluation metrics through a cluster bootstrap procedure with 500 replications, where the clusters are the ICUs. The code for the analyses is available on \href{https://github.com/IYdK/VAKI_diagnosis_ML4H}{GitHub}.

\section{Results}
\label{sec::results}

Of the 176,489 ICU admissions in our data set, 1,638 ICU admissions meet our eligibility criteria. Of these 1,638 admissions, 1,474 are assigned to the train set and 164 to the test set. See \appendixref{apd:fourth} for the baseline characteristics. AKI incidences in the vancomycin and alternative treatment groups are 21.8 and 14.5\%, respectively. 

The distribution of the PC\textsubscript{low} predictions by the BART and LR models are similar, with about 25\% of the ICU admissions assigned a PC\textsubscript{low} of zero and about 50\% of the ICU admissions assigned a PC\textsubscript{low} between 0.5 and 1. In contrast, the distribution of the PC\textsubscript{low} predictions of the RF T-learner is skewed towards high PC\textsubscript{low} estimates (\figureref{fig:pc}).

On average, vancomycin confers an increase in AKI risk compared to the alternative antibiotics in patients that were treated with vancomycin. However, the CIs include the null hypothesis (no average causal effect) in all estimates, except in that of the RF T-learner (\appendixref{apd:fifth}). 

The BART T-learner shows the highest AUC in the evaluation of the factual AKI predictions, followed by the RF and LR models (\appendixref{apd:sixth}). The test set contains 15 patients that were exposed to vancomycin and subsequently developed AKI. The evaluation of the counterfactual PC\textsubscript{low} predictions reveals a relatively low agreement with the medical expert for all T-learner models (\tableref{tab:eval}). Overall, the LR T-learner shows the best agreement.

The overlap in the predicted propensity score distributions between the two treatment groups is relatively high for the BART and LR propensity models, but relatively low for the RF model (\appendixref{apd:seventh}). Lastly, agreement on the PC\textsubscript{low} predictions is relatively high between the BART and LR T-learners, but relatively low between the RF T-learner and the latter (\appendixref{apd:eighth}).

\begin{table}[bp]
\centering
\scriptsize
\floatconts
  {tab:eval}%
  {\caption{Evaluation metrics of the comparison between the PC\textsubscript{low} predictions and the assessments by the medical expert}}%
  {\begin{tabular}{llll}
  \toprule
  Model & AUC                        & PPV       & MSE \\
  \midrule
  BART  & 0.57 (0.32 - 0.76)         & 0.40 (0.25 - 0.67)       & 0.17 (0.10 - 0.28) \\
  RF    & 0.62 (0.42 - 0.87)         & 0.41 (0.22 - 0.67)       & 0.18 (0.12 - 0.29) \\
  LR    & 0.62 (0.38 - 0.81)         & 0.43 (0.25 - 0.67)       & 0.15 (0.10 - 0.28) \\
  \bottomrule
  \end{tabular}}
\end{table}

\section{Discussion}
\label{sec::discussion}

Using a combination of the PC\textsubscript{low} estimand, the TTE framework and ITE estimation, we present a novel data-driven approach for ADE diagnosis to monitor medication safety.

The estimates of all three T-learners show a relatively low agreement with the assessments by the medical expert. Three important limitations of our method became apparent after consulting with the medical expert. First, supratherapeutic vancomycin blood levels are an important factor suggestive of VAKI. Second, an improvement of the AKI marker (serum creatinine, SCr) \textit{after} discontinuation of vancomycin is also an important suggestive factor. Third, an upward trend in SCr before vancomycin initiation may signal a pre-existing AKI. While the first two limitations are not (directly) addressable by our method as they concern post-initiation information, the third limitation could be addressed by adding more information on pre-initiation SCr trends in constructed variables. Providing more information about such variables to the models may improve the individualization of the PC\textsubscript{low} estimates.

Our relatively stringent inclusion and exclusion criteria severely restrict the number of included ICU admissions. Although most exclusions are due to \textit{not} initiating vancomycin or an alternative antibiotic, our cohort of 1,638 ICU admissions comprises of roughly 10\% of the population of interest. It is important to take into account the baseline characteristics of our cohort when aiming to generalize the estimated ATT to different populations. The estimated ATT would likely not be transportable to populations with different distributions of effect modifying patient characteristics.

Inspection of the propensity score distributions and the agreement between the T-learner PC\textsubscript{low} predictions reveal that RF may be inferior in this use-case, possibly due to overfitting and inadequate calibration. The risk of overfitting RF propensity score models has been reported previously, and may be higher in smaller sample size settings \citep{ferri2020propensity}. Inadequate calibration may especially be a risk for RF models \citep{gutman2022propensity}. The more extreme propensity scores predicted by the RF propensity model may result in biased estimates \citep{zhu2021core, zhou2020propensity}. Careful assessment of the causal inference assumptions is of critical importance in the assessment of the models and the identification of optimal modeling techniques. As no gold standard is available for estimated ITEs in real-world settings, standard discrimination and calibration measures of \textit{factual} outcome predictions are not sufficient for assessing model performance.

It is important to note that the goal of causal model-based ADE diagnosis is \textit{not} the emulation of the current best practice, given the known risk of bias associated with human assessors. Instead, the aim is to reliably estimate the individualized PC\textsubscript{low} to aid the human assessor. The current best practice approach does - however - provide the best available comparator. An important limitation of this comparison is the arbitrary mapping of the qualitative causality assessment outcome categories to probabilities.

Future work may address the limitations of our study, including the ones we describe above, the relatively small sample size and our limited capability to emulate the defined hypothetical trial (e.g. by improving the identification of eligible patients). Furthermore, more work is needed on the identification of optimal modeling strategies (e.g. base learner choices) and methods to more objectively compare the qualitative outcomes of causality assessments to model based PC\textsubscript{low} estimates. Importantly, addressing these issues may enable the development of accurate, \textit{data-driven} causal prediction models capable of supporting ADE diagnosis, thereby providing the needed improvement in medication safety monitoring.

\bibliography{jmlr-sample}

\appendix

\section{Assumptions}\label{apd:first}

We let $\vec{Y^a}$ be the counterfactual outcome that would have been observed under treatment $\vec{A=a}$. The assumptions are as follows:

{\bf The stable unit treatment value assumption.} There are no multiple versions of the treatment within treatment indicators that could lead to different outcomes, and an individual's outcome under treatment $\vec{a}$ does not depend on the treatment assignment of other individuals. This assumption also includes the consistency assumption; the counterfactual outcome under an observed treatment $\vec{a}$ is equal to the observed outcome under that treatment: $\vec{A}=\vec{a}\Rightarrow \vec{Y^a}=\vec{Y}$ \citep{vanderweele2013causal, rubin1980randomization}. 

{\bf Weak positivity.} Conditional on the covariates, the probability to receive the alternative option is nonzero: $P(\vec{A}=0|\vec{X})>0$ \citep{imbens2004nonparametric}.

\newcommand{\indep}{\perp \! \! \! \perp}
{\bf Conditional ignorability for controls.} Conditional on the covariates, for ICU admissions that received vancomycin, the counterfactual outcome under the alternative treatment is equal to the factual outcome of ICU admissions treated with the alternative treatment. In other words, the counterfactual outcome under the alternative treatment is independent of the treatment conditional on the covariates: $\vec{Y^0}\indep\vec{A}|\vec{X}$. Note that this is a weaker version of the `regular' conditional ignorability assumption, where one would assume the joint independence of the counterfactual outcomes and the treatment: $\vec{Y^a}\indep\vec{A}|\vec{X}$ \citep{imbens2004nonparametric, heckman1997matching}.

{\bf Monotonicity.} Conditional on the covariates, the relationship between the treatment and the outcome is monotonic. Additionally, vancomycin never prevents censoring events or competing risks compared to the alternative arm \citep{robins1989probability, suzuki2012relations}.

Violations of one or multiple of the above assumptions would lead to upward or downward biased PC\textsubscript{low} estimates. If the conditional monotonicity assumption regarding the relationship between the treatment and the outcome is violated, the PC\textsubscript{low} estimates provide a \textit{worse} lower bound as excess cases are compensated by prevented cases \citep{suzuki2012relations, greenland1988conceptual, greenland1999relation}. If vancomycin prevents censoring events or competing risks compared to the alternative arm, the PC\textsubscript{low} may no longer be a lower bound and instead overestimate the PC. For example, if the alternative antibiotics would instantly lead to death for nearly all patients and vancomycin would not, nearly all AKI cases in the vancomycin arm would be excess cases, and the ERR would approach 1.

\onecolumn
\section{Target Trial Emulation protocol}\label{apd:second}

\begin{table}[H]
\centering
\scriptsize
\rotatebox{90}{
\begin{tabular}{>{\hspace{0pt}}m{0.08\linewidth}>{\hspace{0pt}}m{0.35\linewidth}>{\hspace{0pt}}m{0.35\linewidth}>{\hspace{0pt}}m{0.35\linewidth}}
\toprule
\textbf{Protocol component}  & \textbf{Hypothetical trial}                                                                                                                                                                                                                                                                                                                                                                                                                                                                        & \textbf{Main limitations of our observational data}                                                                                                                                                                                                                                                                                                                                                                                                                                                                        & \textbf{Mitigations of limitations}                                                                                                                                                                                                                                                                                                                                                                                                                                                                                                                                                                                                                                         \\ \hline
Eligibility criteria         & Adult, non-dialysis dependent AKI-free ICU admissions with a confirmed bacterial infection that is treatable with vancomycin or one of the following alternative antibiotics: Clindamycin, linezolid, teicoplanin, meropenem, cefazolin, daptomycin or metronidazole.                                                                                                                                                                                                                              & No data on infection type and susceptibility of bacteria to vancomycin or alternatives. Very few admissions received linezolid, teicoplanin or daptomycin. Limited data on pre-admission AKI. Number of admissions in ICU decreases over time, so lower counts of vancomycin and alternative initiations as length of stay increases.                                                                                                                                                                                      & Use different eligibility criteria: Adult, non-dialysis dependent AKI-free ICU admissions. In daily clinical practice, antibiotics are often started empirically, which means that the infection type and susceptibility of the bacteria is not yet known (it often takes 24 to 48 hours before this is known).~                                                                                                                                                                                                                                                                                                                                                                                                                                                                       \\ \hline
Exclusion criteria           &                                                                                                                                                                                                                                                                                                                                                                                                                                                                                                    &                                                                                                                                                                                                                                                                                                                                                                                                                                                                                                                            & Exclude admission if another treatment option is initiated after initiation of the main treatment. Exclude admission if treatment initiated in first 24 hours or after first 7 days of the ICU admission (see ‘Treatment strategies’, ‘Assignment procedures’ and ‘Eligibility criteria’ for motivations).                                                                                                                                                                                                                                                                                                                                                                                                           \\ \hline
Treatment strategies         & Initiate vancomycin or one of the alternatives at baseline (when eligibility criteria are met), and follow standard therapy procedures for the respective initiated antibiotic (e.g. until 48 hours post negative blood culture).                                                                                                                                                                                                                                                                  & Some patients switched between the different treatment options. Treatment effect might depend on treatment effect in other admissions (e.g. due to antibiotic resistance). Within-treatment group treatments may not be consistent between similar admissions.~                                                                                                                                                                                                                                                             & Exclude admission if another treatment option is initiated after initiation of the main treatment. Assume stable unit of treatment value (SUTVA) and consistency.                                                                                                                                                                                                                                                                                                                                                                                                                                                                                             \\ \hline
Assignment procedures        & Randomized, non-blinded assignment into vancomycin or alternative treatment arms at baseline.                                                                                                                                                                                                                                                                                                                                                                                                      & Assignment into vancomycin or alternative treatment arm not random, but based on admission characteristics. No data available on prevalent users. Unclear timing of some admission characteristics in the first 24 hours of the ICU admission (e.g. acute heart failure, APACHE IV mortality probability). Limited number of admissions within each arm.                                                                                                                                        & Condition on the admission characteristics before treatment initiation and assume ignorability. Exclude admission if treatment is initiated within the first 24 hours of ICU admission. \\ \hline
Follow-up period             & Follow up starts at initiation of vancomycin or alternative (at baseline) and ends at AKI diagnosis, loss to follow-up (i.e. discharge), death or 14 days after treatment initiation, whichever occurs first.                                                                                                                                                                                                                                                                                      & Serum creatinine (SCr) is a lagged indicator of AKI. AKI according to this indicator shortly after treatment initiation is likely not the result of the treatment. In the hypothetical trial this is not a problem given appropriate randomization, because baseline rates of AKI would then be equal across treatment arms.                                                                                                                                                                                                &                                                                                                                                                                                                                                                                                                                                                                                                                                                                                                                                                                                                                                                                            \\ \hline
Outcome                      & AKI during ICU admission according to the KDIGO SCr or urine output (UO) criteria.                                                                                                                                                                                                                                                                                                                                                                                                                 & Relative SCr increase should be calculated using a historical SCr baseline, but not available in our data. UO data is incomplete, heterogeneous across ICUs and lacks resolution for applying KDIGO criteria (daily totals instead of 6 hour totals). Note that these limitations are common in studies utilizing retrospective EHR data \citep{carrero2023defining}.                                                                                                                                                                                                                                                           & Use the first SCr in the first 24 hours of the ICU admission as the SCr baseline. Apply the UO data for the inclusion criteria to aid the recognition of pre-existing AKIs, but do not apply it to diagnose AKI during follow up as the data is too unreliable.~                                                                                                                                                                                                                                                                                                                                                                                                           \\ \hline
Causal contrasts of interest & Per protocol effect for vancomycin-treated cases with AKI conditioned on each admission’s characteristics; Conditional probability of AKI under factual vancomycin treatment ($\mu_1$) vs conditional probability of AKI under counterfactual alternative treatment ($\mu_0$). From these two probabilities we calculate the excess-risk-ratio, which provides a lower bound of the probability of causation (PC\textsubscript{low}). & Positivity must be assessed to ascertain a nonzero probability of receiving an alternative treatment option. Additionally, for patients treated with vancomycin, there may not have been an effective treatment \textit{among the alternative antibiotics}.                                                                                                                                                                                                                                                                & Assume weak positivity. Assume the alternative antibiotics include an effective alternative treatment.                                                                                                                                                                                                                                                                                                                                                                                                                                                                                                                                           \\ \hline
Analysis plan                & We estimate $\mu_1$ and $\mu_0$ using the T-learner causal inference approach. We utilize three different base learners: Bayesian additive regression trees, random forests and logistic regression.~                                                                                                                                                                                                                                                                                      & Relations between the treatment and AKI may not be monotonic conditional on the patient characteristics (leading to an underestimation of the PC\textsubscript{low}) and vancomycin might prevent censoring events (discharge) or competing risks (death) (leading to an overestimation of the PC\textsubscript{low}).                                                                                                                                                                                                                                & Assume a monotonic relationship between treatment and AKI conditional on the admission characteristics, and assume vancomycin never prevents censoring events or competing risks.\\ \bottomrule                                                                                                                                                                                 

\end{tabular}}
\end{table}

\twocolumn
\section{Additional methods}\label{apd:third}

This study is part of the project “Towards a leaRning mEdication Safety system in a national network of intensive Care Units—timely detection of adverse drug Events” (\href{https://www.pharmacoinformaticslab.nl/en/rescue-study/}{RESCUE}). All analyses are conducted in R version 4.0.3 \citep{r2013r}.

\subsection{Data}
\label{sec:app_data}

From the NICE registry we obtain demographics, (chronic) comorbidities at ICU admission, Acute Physiology and Chronic Health Evaluation IV (APACHE IV) admission diagnoses and disease severity scores, physiology measurements and sequential organ failure assessment (SOFA) scores. Via direct data extractions from the 15 ICUs we additionally obtain timed-stamped drug administrations, laboratory findings, arterial blood pressure measurements and renal replacement therapy initiations.

\subsection{Identification of potential confounders}
\label{sec:app_conf}

We identify potential confounders using previous literature on AKI \citep{kellum2012kidney, perazella2012drug, cartin2012risk, neugarten2018female, wiedermann2017causal}, and construct variables using our data where needed to optimally represent the confounders. We include the following variables: acute heart failure, burns, graft and transplant surgery, hypoalbuminemia, hypotension, hypovolemia, major surgery, mechanical ventilation, sepsis, trauma, age, sex, alcohol abuse, chronic carviovascular disease, chronic kidney disease, chronic pulmonary disease, diabetes mellitus, chronic liver disease, malignancy, obesity, the APACHE IV predicted mortality probability, serum creatinine (SCr) baseline and estimated glomerular filtration rate (eGFR). We additionally include longitudinally measured pre-baseline patient parameters, including the mean arterial blood pressure (MAP), urine output rate (UO), serum albumin, leukocyte count, temperature and SOFA scores. We summarize longitudinal values of the latter variables to one value by clinically relevant aggregation functions (e.g. the maximum SOFA score, see \appendixref{apd:fourth}). Moreover, we calculate a variable specifying the time between ICU admission and the initiation of treatment. Lastly, we include nephrotoxic co-medications: we identify individual nephrotoxic drugs by their World Health Organization Anatomical Therapeutic Chemical (ATC) code in a recent consensus-based list of nephrotoxins in the ICU setting \citep{gray2022consensus}. We include all drugs associated with a nephrotoxicity rating of at least 0.5 ("possible"). For each of these nephrotoxins, we create a variable that indicates whether the ICU admission received the drug or not before or at baseline. In total, we include 107 variables, of which 77 are nephrotoxic drugs.

\subsection{Missing data imputation}
\label{sec:app_miss}

Missing data are imputed with the mean or median for continuous or categorical data, respectively. We do not impute the outcome.

\subsection{Variable selection}
\label{sec:app_vars}

We apply automated variable selection during each bootstrap iteration. First, binary variables with a prevalence of less than 2\% are removed \citep{patrick2011implications}. Second, we remove the longitudinally measured parameters and nephrotoxins with a univariable association with AKI with $p>0.2$ \citep{brookhart2006variable}.

\subsection{Model recalibration}
\label{sec:app_cal}

The propensity score models and T-learner base learners are all recalibrated on the training data during each bootstrap iteration. We apply the recently developed GUESS recalibration algorithm \citep{schwarz2019guess}.

\subsection{Funding}
\label{sec:app_fund}

This study was funded by The Netherlands Organization for Health Research and Development (ZonMw), project number \href{https://projecten.zonmw.nl/nl/project/towards-learning-medication-safety-system-national-network-intensive-care-units-timely}{848018004}. ZonMw had no role in the design of the study, the collection, analysis and interpretation of the data, or in writing the manuscript.

\subsection{Ethics approval and patient consent}
\label{sec:app_eth}

This study was exempted from requiring ethics approval (waiver W19\_433 \# 19.499) on 14 November 2019 by the Medical Ethics Committee of the Amsterdam University Medical Centers, location University of Amsterdam, The Netherlands, as it did not fall within the scope of the Dutch Medical Research Involving Human Subjects Act (WMO). Due to the large number of patients in our data, approaching all individuals to obtain informed consent would require an unreasonable effort. Under Dutch law, this study was therefore exempted from the requirement to obtain informed consent.

\subsection{Data availability}
\label{sec:app_share}

The data is not publicly available due to the data sharing agreements with the participating ICUs. Access to the data might only be provided after explicit consent from each separate participating ICU. For conditions under which access is allowed, please see: \url{https://uvaauas.figshare.com/articles/dataset/RESCUE_Metadata/22309996}.

\onecolumn
\section{Baseline characteristics}\label{apd:fourth}
\begin{table}[H]
\captionsetup{singlelinecheck = false} %
\caption*{Table D.1: Baseline characteristics of the ICU admissions in the train set}%
\centering
\scriptsize
\begin{tabular}{>{\hspace{0pt}}m{0.563\linewidth}>{\hspace{0pt}}m{0.16\linewidth}>{\hspace{0pt}}m{0.16\linewidth}>{\hspace{0pt}}m{0.025\linewidth}}
\toprule
\textbf{Characteristic}                                                               & \textbf{Alternative \newline (n = 856)} & \textbf{Vancomycin \newline (n = 618)} &   \\ \midrule
Age (Years), median (Q1 - Q3)                                                & 63.0 (53.0 - 73.0)    & 66.0 (55.2 - 74.0)   &   \\
Male sex, No. (\%)                                                        & 518 (60.5)            & 374 (60.5)           &   \\
Planned admission, No. (\%)                                                  & 94 (11.0)             & 106 (17.2)           &   \\
Admission type                                                               &                       &                      &   \\
~~~~~
  Medical, No. (\%)                                                    & 630 (73.8)            & 381 (61.7)           &   \\
~~~~~
  Emergency surgical, No. (\%)                                         & 112 (13.1)            & 124 (20.1)           &   \\
~~~~~
  Elective surgical, No. (\%)                                          & 112 (13.1)            & 113 (18.3)           &   \\
APACHE IV score, median (Q1 - Q3)                                            & 52.5 (36.0 - 70.0)    & 55.0 (44.0 - 70.0)   &   \\
APACHE IV mortality probability, median
  (Q1 - Q3)                          & 0.2 (0.1 - 0.4)       & 0.2 (0.1 - 0.4)      &   \\
SCr baseline (mg/dL), median (Q1 - Q3)                                       & 0.8 (0.7 - 1.2)       & 0.9 (0.7 - 1.2)      &   \\
eGFR baseline (mL/min/1.73 m²), median
  (Q1 - Q3)                           & 82.5 (56.4 - 108.9)   & 80.8 (56.4 - 107.7)  &   \\
Highest SCr during admission (g/dL),
  median (Q1 - Q3)                      & 0.9 (0.7 - 1.2)       & 1.0 (0.7 - 1.2)      &   \\
Lowest urine output during admission
  (mL/kg/h), median (Q1 - Q3)           & 0.9 (0.7 - 1.3)       & 0.9 (0.7 - 1.3)      &   \\
Lowest serum albumin during admission
  (g/dL), median (Q1 - Q3)             & 2.5 (2.1 - 2.7)       & 2.4 (1.8 - 2.7)      &   \\
Lowest MAP during admission (mmHg),
  median (Q1 - Q3)                       & 59.0 (51.0 - 62.0)    & 57.0 (50.0 - 62.0)   &   \\
Highest SOFA score during admission,
  median (Q1 - Q3)                      & 7.0 (5.0 - 9.0)       & 8.0 (6.0 - 10.0)     &   \\
Highest temperature during admission
  (°C), median (Q1 - Q3)                & 38.0 (37.9 - 38.8)    & 38.2 (38.0 - 39.3)   &   \\
Highest leukocyte count during admission
  (10**9 cells/L), median (Q1 - Q3) & 15.6 (11.6 - 21.4)    & 16.8 (11.6 - 23.0)   &   \\
Acute AKI risk factors                                                       &                       &                      &   \\
~~~~~
  Acute heart failure, No. (\%)                                        & 114 (13.3)            & 87 (14.1)            &   \\
~~~~~
  Burns, No. (\%)                                                      & 1 (0.1)               & 0 (0.0)              &   \\
~~~~~
  Graft or transplant surgery, No. (\%)                                & 45 (5.3)              & 53 (8.6)             &   \\
~~~~~
  Hypoalbuminemia, No. (\%)                                            & 699 (81.7)            & 525 (85.0)           &   \\
~~~~~
  Hypotension, No. (\%)                                                & 223 (26.1)            & 201 (32.5)           &   \\
~~~~~
  Hypovolemia, No. (\%)                                                & 22 (2.6)              & 13 (2.1)             &   \\
~~~~~
  Major surgery, No. (\%)                                              & 209 (24.4)            & 198 (32.0)           &   \\
~~~~~
  Mechanical ventilation, No. (\%)                                     & 542 (63.3)            & 470 (76.1)           &   \\
~~~~~
  Sepsis - admission diagnosis, No. (\%)                               & 112 (13.1)            & 93 (15.0)            &   \\
~~~~~
  Sepsis - longitudinal, No. (\%)                                      & 273 (31.9)            & 239 (38.7)           &   \\
~~~~~
  Trauma, No. (\%)                                                     & 97 (11.3)             & 14 (2.3)             &   \\
Chronic AKI risk factors                                                     &                       &                      &   \\
~~~~~
  Alcohol abuse, No. (\%)                                              & 191 (22.3)            & 141 (22.8)           &   \\
~~~~~
  Cardiovascular disease, No. (\%)                                     & 232 (27.1)            & 161 (26.1)           &   \\
~~~~~
  Chronic kidney disease, No. (\%)                                     & 18 (2.1)              & 19 (3.1)             &   \\
~~~~~
  Chronic pulmonary disease, No. (\%)                                  & 154 (18.0)            & 78 (12.6)            &   \\
~~~~~
  Diabetes mellitus, No. (\%)                                          & 135 (15.8)            & 97 (15.7)            &   \\
~~~~~
  Liver disease, No. (\%)                                              & 14 (1.6)              & 10 (1.6)             &   \\
~~~~~
  Malignancy, No. (\%)                                                 & 115 (13.4)            & 104 (16.8)           &   \\
~~~~~
  Obesity, No. (\%)                                                    & 114 (13.3)            & 103 (16.7)           &   \\
Nephrotoxin exposure                                                         &                       &                      &   \\
~~~~~
  J01GB03 - Gentamicin, No. (\%)                                       & 123 (14.4)            & 95 (15.4)            &   \\
~~~~~
  J05AB01 - Aciclovir, No. (\%)                                        & 11 (1.3)              & 22 (3.6)             &   \\
~~~~~
  J01DC02 - Cefuroxime, No. (\%)                                       & 55 (6.4)              & 109 (17.6)           &   \\
~~~~~
  J01DD02 - Ceftazidime, No. (\%)                                      & 25 (2.9)              & 45 (7.3)             &   \\
~~~~~
  J01MA02 - Ciprofloxacin, No. (\%)                                    & 131 (15.3)            & 160 (25.9)           &   \\
~~~~~
  J01CR02 - Amoxicillin And Beta-Lactamase Inhibitor, No. (\%)         & 84 (9.8)              & 42 (6.8)             &   \\
~~~~~
  J01CR05 - Piperacillin And Beta-Lactamase Inhibitor, No. (\%)        & 45 (5.3)              & 40 (6.5)             &   \\
~~~~~
  S01EC01 - Acetazolamide, No. (\%)                                    & 37 (4.3)              & 33 (5.3)             &   \\
~~~~~
  C03CA01 - Furosemide, No. (\%)                                       & 284 (33.2)            & 342 (55.3)           &   \\
~~~~~
  N01AH03 - Sufentanil, No. (\%)                                       & 173 (20.2)            & 143 (23.1)           &   \\
~~~~~
  J01CF05 - Flucloxacillin, No. (\%)                                   & 43 (5.0)              & 24 (3.9)             &   \\
~~~~~
  J01CE01 - Benzylpenicillin, No. (\%)                                 & 20 (2.3)              & 15 (2.4)             &   \\
~~~~~
  A02BC05 - Esomeprazole, No. (\%)                                     & 154 (18.0)            & 125 (20.2)           &   \\
~~~~~
  J01EE01 - Sulfamethoxazole And Trimethoprim, No. (\%)                & 28 (3.3)              & 38 (6.1)             &   \\
Time in ICU before initiation (Days),
  median (Q1 - Q3)                     & 1.9 (1.4 - 3.2)       & 2.6 (1.7 - 4.2)      &   \\
AKI, No. (\%)                                                                & 124 (14.5)            & 135 (21.8)           &   \\
AKI stage                                                                    &                       &                      &   \\
~~~~~
  Stage 1, No. (\%)                                                    & 76 (61.3)             & 85 (63.0)            &   \\
~~~~~
  Stage 2, No. (\%)                                                    & 17 (13.7)             & 18 (13.3)            &   \\
~~~~~
  Stage 3, No. (\%)                                                    & 31 (25.0)             & 32 (23.7)            &   \\
~~~~~
  RRT, No. (\%)                                                        & 26 (21.0)             & 23 (17.0)            &   \\
Length of stay (Days), median (Q1 - Q3)                                      & 5.9 (3.4 - 10.2)      & 8.1 (4.8 - 16.1)     &   \\
ICU mortality, No. (\%)                                                      & 101 (11.8)            & 93 (15.0)            &   \\
Hospital mortality, No. (\%)                                                 & 149 (17.4)            & 138 (22.3)           &  
\end{tabular}
\end{table}

\newpage
\section{ATT estimates}\label{apd:fifth}

\begin{table}[H]
\captionsetup{singlelinecheck = false} %
\caption*{Table E.1: ATT estimates of the T-learner and IPTW approaches}%
\centering
 \floatconts
  {tab:att}%
  {}%
  {\begin{tabular}{lllll}
  \toprule
        & AKI risk alternative & AKI risk vancomycin & ARD & RR         \\
        \midrule
  Unadjusted       & 0.15 (0.12 - 0.20)     & 0.22 (0.15 - 0.28)    & 0.07 (0.01 - 0.12)       & 1.44 (1.07 - 1.83) \\
  T-learner - BART & 0.19 (0.14 - 0.26)     & 0.22 (0.15 - 0.28)    & 0.03 (-0.02 - 0.08)      & 1.16 (0.90 - 1.48) \\
  T-learner - RF   & 0.16 (0.11 - 0.24)     & 0.22 (0.15 - 0.28)    & 0.06 (0.00 - 0.10)       & 1.34 (1.03 - 1.77) \\
  T-learner - LR   & 0.18 (0.14 - 0.26)     & 0.22 (0.16 - 0.32)    & 0.03 (-0.01 - 0.12)      & 1.19 (0.94 - 1.68) \\
  IPTW - BART      & 0.19 (0.14 - 0.27)     & 0.22 (0.15 - 0.28)    & 0.03 (-0.04 - 0.08)      & 1.14 (0.85 - 1.52) \\
  IPTW - RF        & 0.19 (0.12 - 0.45)     & 0.22 (0.15 - 0.28)    & 0.03 (-0.23 - 0.09)      & 1.17 (0.59 - 2.27) \\
  IPTW - LR        & 0.18 (0.13 - 0.25)     & 0.22 (0.15 - 0.28)    & 0.03 (-0.02 - 0.09)      & 1.18 (0.91 - 1.55) \\
  \bottomrule
  \end{tabular}}
\end{table}

\section{Factual AKI prediction evaluation}\label{apd:sixth}

\begin{table}[H]
\captionsetup{singlelinecheck = off} %
\caption*{Table F.1: Evaluation of the factual AKI\newline predictions by the T-learner base models}%
\begin{tabular}{ll}
\toprule
T-learner base model              & AUC           \\
                \midrule
Vancomycin - BART & 0.75 (0.52 - 0.86) \\
Vancomycin - RF   & 0.72 (0.59 - 0.84) \\
Vancomycin - LR   & 0.71 (0.50 - 0.84) \\
Alternative - BART    & 0.74 (0.59 - 0.82) \\
Alternative - RF      & 0.73 (0.62 - 0.83) \\
Alternative - LR      & 0.72 (0.51 - 0.81) \\  
\bottomrule
\end{tabular}
\end{table}

\section{Propensity score distributions}\label{apd:seventh}

\begin{figure}[H]
\floatconts
  {fig:prop}
  {}
  {\includegraphics[scale=0.55]{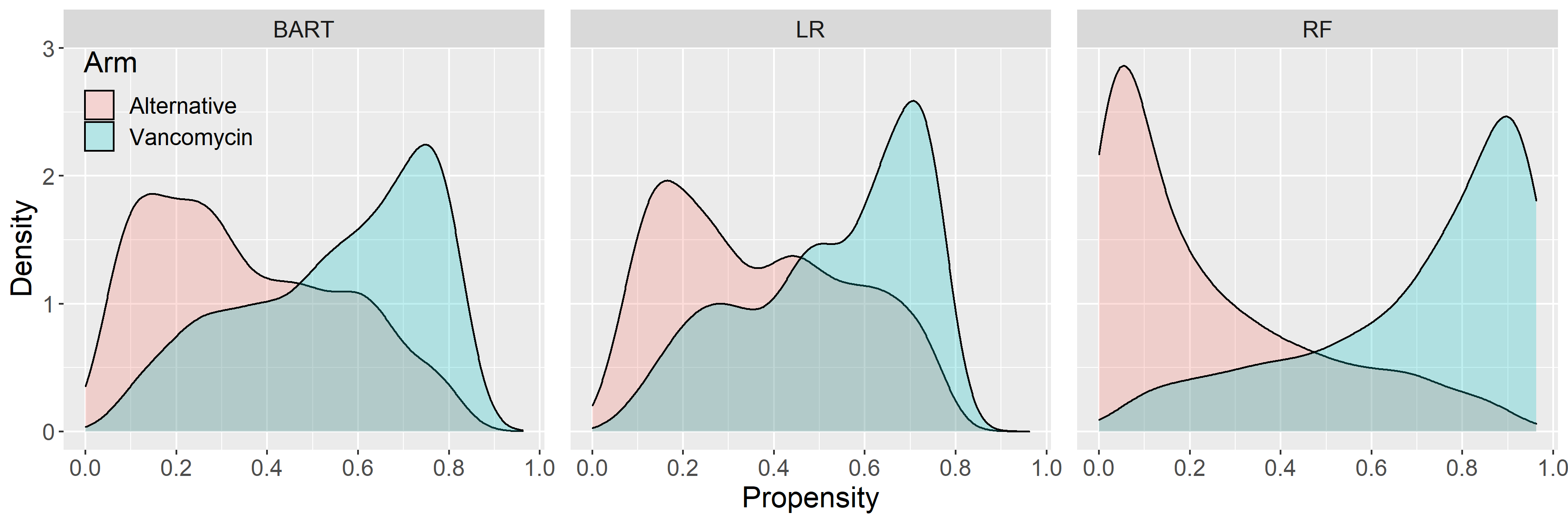}}
  \captionsetup{singlelinecheck = false} %
\caption*{Figure G.1: Distributions of the propensity scores predicted by each propensity score model, stratified by treatment arm}%
\end{figure}

\newpage
\section{Correlations between the T-learners' predictions}\label{apd:eighth}

\begin{table}[H]
\captionsetup{singlelinecheck = false} %
\caption*{Table H.1: Correlations between the T-learners' PC\textsubscript{low}\newline predictions for the test set’s ICU admissions that\newline received vancomycin and subsequently developed AKI\newline (Pearson correlation coefficient)}%
{\begin{tabular}{llll}
\toprule
     & BART                        & RF       & LR \\
  \midrule
  BART  & 1         &         &   \\
  RF    & 0.55 (0.10 - 0.86)         & 1       &   \\
  LR    & 0.71 (0.19 - 0.93)         & 0.44 (-0.08 - 0.78)       & 1 \\
  \bottomrule
  \end{tabular}}
\end{table}

\end{document}